\documentclass[a4paper,twoside]{article}
\pdfoutput=1
\usepackage{epsfig}
\usepackage{subcaption}
\usepackage{calc}
\usepackage{amssymb}
\usepackage{amstext}
\usepackage{amsmath}
\usepackage{amsthm}
\usepackage{multicol}
\usepackage{pslatex}
\usepackage{apalike}
\usepackage{algorithm2e}
\usepackage[bottom]{footmisc}
\usepackage{url}
\usepackage{SCITEPRESS}    
\usepackage{amsmath} 

\begin{document}

\title{FEST: A Unified Framework for Evaluating Synthetic Tabular Data}

\author{\authorname{Weijie Niu, Alberto Huertas Celdran, Karoline Siarsky, Burkhard Stiller}
\affiliation{Communication Systems Group CSG, Department of Informatics, University of Zurich UZH, CH--8050 Zürich, Switzerland}
\email{\{niu, huertas, stiller\}@ifi.uzh.ch, karoline.siarsky@uzh.ch}
}


\abstract{Synthetic data generation, leveraging generative machine learning techniques, offers a promising approach to mitigating privacy concerns associated with real-world data usage. Synthetic data closely resemble real-world data while maintaining strong privacy guarantees. However, a comprehensive assessment framework is still missing in the evaluation of synthetic data generation, especially when considering the balance between privacy preservation and data utility in synthetic data. This research bridges this gap by proposing FEST, a systematic framework for evaluating synthetic tabular data. FEST integrates diverse privacy metrics (attack-based and distance-based), along with similarity and machine learning utility metrics, to provide a holistic assessment. We develop FEST as an open-source Python-based library and validate it on multiple datasets, demonstrating its effectiveness in analyzing the privacy-utility trade-off of different synthetic data generation models. The source code of FEST is available on Github.}

\onecolumn \maketitle \normalsize \setcounter{footnote}{0} \vfill

\section{\uppercase{Introduction}}
\label{sec:introduction}
The rapid growth of Machine Learning (ML) and Artificial Intelligence (AI) is driving significant changes in industries like healthcare, finance, and education, improving decision-making and making operations more efficient. However, the successful implementation of these AI technologies depends on access to large-scale high-quality datasets which are essential for training complex models. That need for large datasets comes with several challenges. Firstly, data scarcity is an issue in certain domains. It can be challenging to collect sufficient and high-quality data, especially when missing or incomplete data can affect the accuracy of AI algorithms. Secondly, collecting large amounts of data can be expensive, especially when dealing with sensitive information. Most importantly, which is the main driving point of this work, there are significant concerns related to the privacy and sensitivity of data, such as healthcare or financial records, when used to train generative AI algorithms. 

To address these challenges, synthetic data generation has been recognized as an effective solution \cite{kotelnikov2023tabddpm}. By generating artificial data that mimics the statistical properties of real-world data, synthetic data offers the potential to support downstream tasks, such as training AI models without compromising individuals' privacy. It is often considered to preserve privacy \cite{zhao2024ctab} technically better. However, the reality is more nuanced. Synthetic data can still contain personally identifiable information about individuals from the original data and may not achieve complete privacy protection, so the use of synthetic data remains controversial. While privacy-preserving techniques are crucial, it is also important to ensure that the synthetic data remains useful for its intended applications. So, finding a balance between privacy and data utility is essential. There is a significant research gap on the extent to which synthetic data can effectively preserve privacy in combination with data utility.







This paper addresses a comprehensive framework to assess the privacy and utility of synthetic data generated by various methods, including Generative Adversarial Networks (GAN), Variational Autoencoders (VAE), and other machine learning techniques. The main contributions of this paper include: 
\begin{itemize}
    \item Developing a framework FEST (Framework for Evaluating Synthetic Tabular Data) that integrates a diverse set of privacy metrics, encompassing both attack-based and distance-based measures, to provide a holistic evaluation of privacy risks. The framework also includes statistical similarity assessment and machine learning utility assessment. The FEST is open-sourced \cite{FEST} for researchers in the community to use in evaluating synthetic data generation models.
    \item Implementing and evaluating the framework on multiple real-world datasets to demonstrate its effectiveness in assessing the privacy-utility trade-off of different synthetic data generation models.
    \item Providing insights into the strengths and limitations of various privacy metrics and their applicability to different types of synthetic data. 
\end{itemize}

\section{\uppercase{Related Work}}

The generation and evaluation of synthetic data, particularly for privacy-preserving purposes, is an area of increasing importance \cite{sanchezprivacy}. Existing research has explored various approaches, including GANs \cite{Yale2020GenerationData} and VAEs \cite{Lu2023MachineReview}. However, concerns remain regarding the effectiveness of these methods in ensuring robust privacy protection. Studies including \cite{StadlerSyntheticMirage} have challenged the assumption that synthetic data inherently guarantees privacy, highlighting the potential for re-identification and inference attacks. This emphasizes the need for rigorous evaluation frameworks to assess the true privacy risks associated with synthetic data. 

This gap is further highlighted by comprehensive surveys including \cite{Bauer2024ComprehensiveSurvey}, which analyze numerous synthetic data generation methods but may not provide a standardized and holistic evaluation framework. In the context of specific domains, such as healthcare, research efforts have focused on applying and evaluating synthetic data generation methods \cite{HernandezSyntheticReview}. 

Many existing studies rely mainly on limited privacy metrics, such as Distance of Closest Record (DCR) and Nearest Neighbor Distance Ratio (NNDR) \cite{xu2019modeling}, \cite{zhao2021ctab}, \cite{kotelnikov2023tabddpm}, questioning the effectiveness in evaluating privacy in various contexts. Some papers have proposed new metrics. For instance, the study\cite{Raab2024PracticalData} have introduced valuable metrics such as DiSCO and repU, while others have emphasized the importance of Singling Out, Linkability, and Inference risks \cite{giomi2022unified}. Yet, they come with limitations, such as focusing on specific types of metrics or neglecting considerations of data utility.  


\section{PRELIMINARIES}
\label{chap: gen model}
A key component of this research is based on the understanding of synthetic data generation methods. To thoroughly validate our framework and ensure its robustness across different data generation techniques, this study implements and evaluates the following models: Conditional Tabular GAN (CTGAN), Gaussian Mixture (GM), Gaussian Copula (GC), CopulaGAN, TVAE, and a Random Model. By applying these models to real-world datasets and assessing the privacy and utility of the resulting synthetic data, this study can gain insights into the strengths and limitations of our framework in diverse scenarios. For each model, a synthetic dataset of the same size as the original one is generated.  

\subsection{Conditional Tabular GAN (CTGAN)}
CTGAN \cite{xu2019modeling} is a specialized GAN designed for synthesizing tabular data, and handling mixed data types and imbalanced datasets using conditional vectors. It employs a generator to create synthetic samples and a discriminator to distinguish them from real data. The iterative training process refines both components until the synthetic data closely resembles the real data.

\subsection{Gaussian Mixture (GM)}
Gaussian Mixture Models (GM) are statistical models representing data as a mixture of Gaussian distributions, each reflecting different subgroups within the dataset. The model estimates the means, variances, and mixing proportions using techniques like the Expectation-Maximization (EM) algorithm. GM generates synthetic data reflecting the original data's structure, particularly useful for data with multiple peaks or clusters. It is included as a baseline comparison to more advanced models.

\subsection{Gaussian Copula (GC)}
The Gaussian Copula (GC) model is a statistical method used for generating synthetic data that preserves the correlation structure between variables. It transforms data into a uniform distribution using the Cumulative Distribution Function (CDF) and applies a Gaussian copula to model dependencies. This ensures the synthetic data maintains the original correlation patterns. GC is included as another statistical baseline for comparison.

\subsection{CopulaGAN}
CopulaGAN, a variation of CTGAN, enhances performance by incorporating a CDF-based transformation, similar to GC. This allows CopulaGAN to better capture complex dependencies and generate more realistic synthetic data by leveraging both statistical and GAN-based techniques.

\subsection{Tabular Variational Autoencoder (TVAE)}
TVAE \cite{xu2019modeling} is a Variational Autoencoder designed for tabular data. It learns a probabilistic representation of the data, encoding it into a lower-dimensional latent space and then decoding it to generate new synthetic data points. This method allows TVAE to preserve key statistical properties and relationships of the original data.

\subsection{Random Model}
The Random Model serves as a baseline, randomly sampling data points from the original dataset with or without replacement. With replacement, a data point can be chosen multiple times; without replacement, it's chosen only once, limiting the synthetic dataset to the original size.

\section{METHODOLOGY}
\begin{figure*}
    \centering
    \includegraphics[width=1\textwidth]{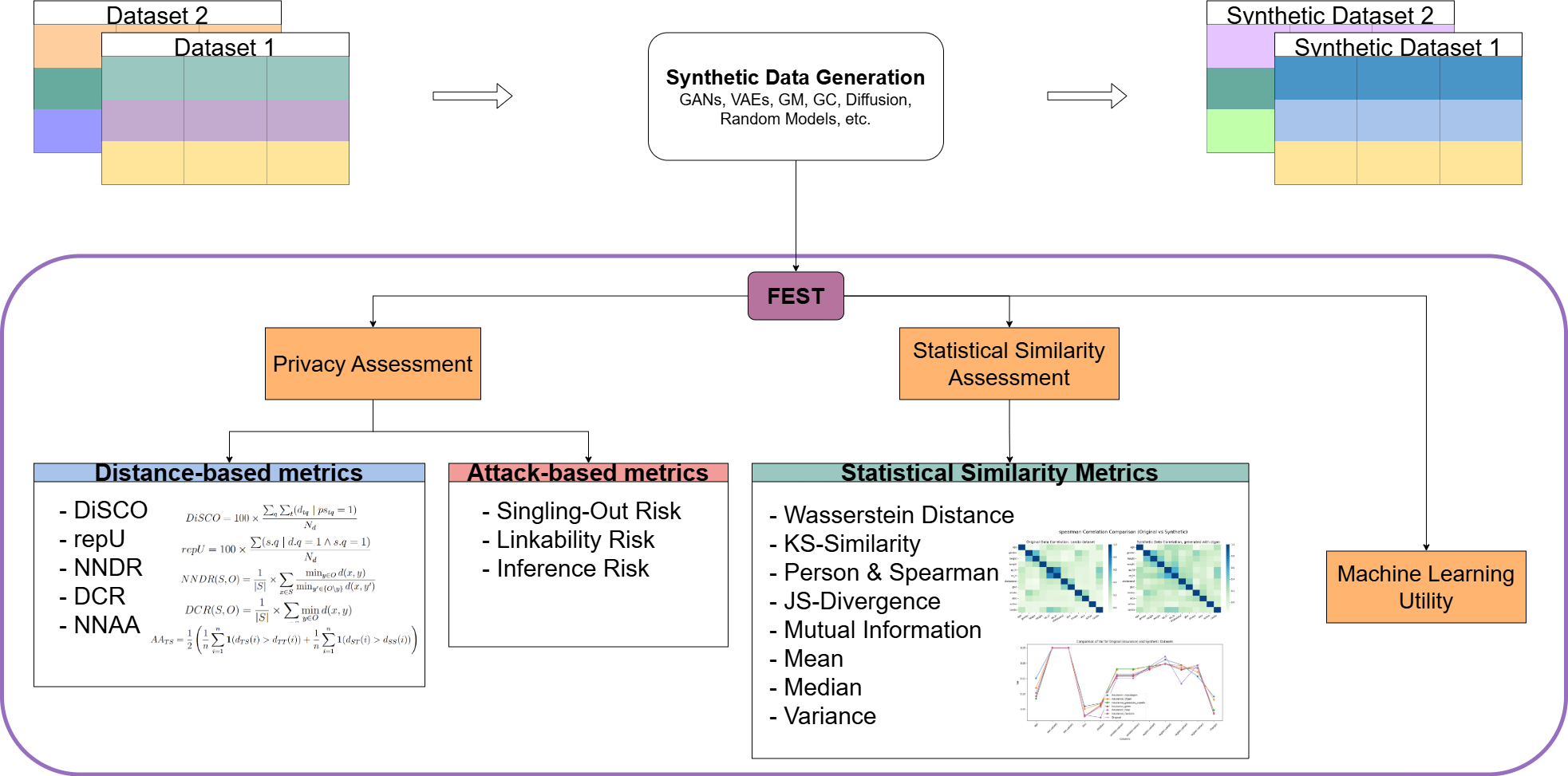}    
    \caption{Tree-Based Visualization of the Framework}
    \label{fig:framework}
\end{figure*}

This chapter introduces the proposed solution FEST, a unified framework for the evaluation of synthetic tabular data generation, encompassing three key pillars: Privacy Assessment, Statistical Similarity Assessment, and Machine Learning Utility. The FEST framework is illustrated in Figure \ref{fig:framework}. 

In this framework, the privacy of synthetic data will be evaluated using different kinds of metrics. This study categorizes the privacy metrics into Attack-based metrics and Distance-based metrics. These include singling out risk, linkability risk, and inference risk metrics, which help identify how vulnerable individuals are to re-identification or attribute inference from synthetic data. Distance-based metrics such as Disclosive in Synthetic Correct Original (DiSCO), Replicated Uniques (repU), Nearest-Neighbor Distance Ratio (NNDR), Distance of Closest Record (DCR) and Nearest-Neighbor Adversarial Accuracy (NNAA) will also be considered. These metrics offer detailed insights into how closely synthetic records may resemble real data and the risks associated with synthetic data generation. Each of these metrics will be described and analyzed below. 

Statistical similarity will be measured using various metrics that compare the statistical properties of the synthetic and original datasets. These metrics will be elaborated in the following sections. Meanwhile, machine learning utility, which assesses the ability of synthetic data to support model training, is discussed.

\subsection{Privacy Metrics}

Privacy metrics are subdivided into attack-based and distance-based metrics.

\subsubsection{Attack-based Metrics}
To quantify privacy risks based on attacks, this study includes three attacks, especially based on GDPR guidelines: Singling Out (S-Out), Linkability (Link), and Attribute Inference (Inf). Their implementation follows Anonymeter \cite{giomi2022unified}.

\begin{itemize}
    \item \textbf{Singling Out Risk}
    \label{item_sout}
        Singling out refers to pinpointing a distinct data record from an original dataset using a specific attribute combination. For example, an attacker might identify a unique combination like a 45-year-old data scientist in ZIP code 12345 in the synthetic data. While isolating an individual does not necessarily lead to re-identification, it can enable privacy-compromising attacks \cite{Haque2022SemanticSummarization}.

    \item \textbf{Linkability Risk}
    \label{item_link}
        Linkability Risk refers to the potential for different pieces of information about an individual to be combined in a way that can identify them. An attacker with knowledge of some attributes from one source (Dataset A) and other attributes from another source (Dataset B) could use the synthetic dataset to match these attributes and identify that they belong to the same individual from the original dataset \cite{giomi2022unified}.
        
    \item \textbf{Attribute Inference Risk}
    \label{item_inf}
        Attribute inference attack evaluates the risk that an attacker can deduce sensitive attributes using synthetic data. The attacker, having partial knowledge of certain attributes (auxiliary information, AUX), attempts to infer unknown sensitive attributes of target records by analyzing the synthetic dataset. For example, if an attacker knows the age and ZIP code (AUX) of a target individual from the original dataset, the goal is to infer their medical condition (the sensitive attribute) using the synthetic dataset.
\end{itemize}

\subsubsection{Distance-based Metrics}
Distance-based metrics evaluate privacy by quantifying the similarity (distance) between the original and synthetic data records. 
\begin{itemize}
    \item \textbf{DiSCO:} DiSCO assesses attribute disclosure risk in synthetic data by determining the proportion of records disclosive in both synthetic and original data. A disclosive record can reveal target attributes about individuals when specific quasi-identifiers are analyzed \cite{Raab2024PracticalData}. It indicates how much sensitive information from original data can be inferred from the synthetic data. Quasi-identifiers, while not unique on their own, can be combined to identify individuals, such as age, gender and ZIP code \cite{Motwani2007EfficientQuasi-Identifiers}. The DiSCO metric is calculated as (\ref{eq: disco}):
\begin{equation}
    DiSCO = 100 \times \frac{\sum_{q} \sum_{t}(d_{tq} \mid ps_{tq} = 1)}{N_d}
\label{eq: disco}
\end{equation}

    \begin{itemize}
        \item $q$: Quasi-identifiers
        \item $t$: Target attribute.
        \item $d_{tq}$: Distribution of the target attribute in the original data for a given set of quasi-identifiers. This represents the frequency or count of records that have a specific value of the target attribute t for each combination of quasi-identifiers q.
        \item $ps_{tq}$: Proportion of the target attribute in the synthetic data. This measures the relative frequency or probability that the target attribute t occurs in the synthetic data for each combination of quasi-identifiers q.
        \item $N_d$: Total number of records in original data.
    \end{itemize}
    
    \item \textbf{repU:}
    The repU metric measures the risk of identity disclosure by evaluating the replication of unique quasi-identifier combinations in synthetic data. It checks how many unique records from the original dataset are replicated in synthetic datasets \cite{Raab2024PracticalData}. repU is calculated as (\ref{eq: repu})
    \begin{equation}
    repU = 100 \times \frac{\sum(s_q \mid d_q = 1 \land s_q = 1)}{N_d}
        \label{eq: repu}
    \end{equation}
    \begin{itemize}
        \item $s_q$: Value in the synthetic data for a given set of quasi-identifiers. When $s_q = 1$ it means that the particular combination of quasi-identifiers in the synthetic data is unique.
        \item $d_q$: Value in the original data for a given set of quasi-identifiers. When $d_q = 1$ it means that the particular combination of quasi-identifiers in the original data is unique.
        \item $N_d$: Total number of records in original data.
    \end{itemize}

    \item \textbf{NNDR:} The NNDR is a metric that evaluates the relative distances between the data points. The NNDR is calculated as the ratio of the Euclidean distance between each record in the synthetic dataset and its closest corresponding record in the original dataset to the distance to the second-closest record. Mathematically, for a synthetic dataset S and an original dataset O, the NNDR is calculated as (\ref{eq: nndr}):
\begin{equation}
NNDR(S, O) = \frac{1}{|S|} \times \sum_{x \in S} \frac{\min_{y \in O} d(x, y)}{\min_{y’ \in \{O \setminus {y}\}} d(x, y’)}
\label{eq: nndr}
\end{equation}
In this formula:

\begin{itemize}
    \item $d(x, y)$ represents the Euclidean distance between records x and y
    \item $\min_{y \in O} d(x, y)$ is the distance to the closest record.
    \item $\min_{y’ \in O \setminus {y}} d(x, y’)$ is the distance to the second closest record.
\end{itemize}

    This ratio helps in identifying how isolated a data point is within the dataset. A lower NNDR indicates that a data point is more isolated, while a higher NNDR suggests it is in a densely populated area of the dataset. The values of the NNDR metric range from 0 to 1.
    \item \textbf{DCR:} The DCR measures how far each synthetic data point is from the nearest data point in the original dataset. This metric is used to assess the similarity between synthetic and original data. A DCR of zero means that the synthetic data point is identical to the original data point, which could pose a privacy risk. Higher DCR values indicate that the synthetic data is more distinct from the original data, reducing the risk of privacy breaches. The value range of DCR is from 0 to infinity. Different types of distance metrics can be used to calculate DCR, such as Euclidean distance (straight-line distance) and Manhattan distance. FEST implements Euclidean distance. Mathematically, for a synthetic dataset $S$ and an original dataset $O$, the DCR is calculated as (\ref{eq: dcr})

\begin{equation}
    DCR(S, O) = \frac{1}{|S|} \times \sum_{x \in S} \min_{y \in O} d(x, y)
\label{eq: dcr}
\end{equation}

\begin{itemize}
    \item $d(x, y)$ represents f.e. the Euclidean distance between records x and y.
    \item $\min_{y \in O} d(x, y)$ is the distance to the closest record.

\end{itemize}

    \item \textbf{NNAA}: Nearest-Neighbor Adversarial Accuracy assesses synthetic data quality while preserving privacy. If synthetic data is indistinguishable from real data, a nearest-neighbor classifier should not differentiate them. Adversarial Accuracy (AA) quantifies how well an adversary can distinguish between real and synthetic data. The NNAA is calculated as (\ref{eq:nnaa}):
    \begin{multline}
    AA_{TS} = \frac{1}{2} \Biggl( \frac{1}{n} \sum_{i=1}^n \mathbf{1}(d_{TS}(i) > d_{TT}(i)) \\
    + \frac{1}{n} \sum_{i=1}^n \mathbf{1}(d_{ST}(i) > d_{SS}(i)) \Biggr)
    \label{eq:nnaa}
\end{multline}

    \begin{itemize}
        \item $d_{TS}(i)$: The distance between the i-th record in the target dataset (T) and its nearest neighbor in the source dataset (S). This measures how close an original data point is to its nearest synthetic counterpart.
        \item $d_{ST}(i)$: The distance between the i-th record in the source dataset (S) and its nearest neighbor in the target dataset (T). This measures how close a synthetic data point is to its nearest original counterpart.
        \item $d_{TT}(i)$: The distance between the i-th record in the target dataset (T) and its nearest neighbor within the target dataset (excluding itself). This serves as a baseline for comparison within the original data.
        \item $d_{SS}(i)$: The distance between the i-th record in the source dataset (S) and its nearest neighbor within the source dataset (excluding itself). This serves as a baseline for comparison within the synthetic data.
        \item $\mathbf{1}$: An indicator function that takes the value 1 if the condition inside the parentheses is true, and 0 otherwise.
        \item $n$: The total number of records.
    \end{itemize}

    An AA score close to 0.5 means that the synthetic data is indistinguishable from the original data, indicating a good balance of resemblance and privacy. A higher than 0.5 AA score suggests that the synthetic data can be easily distinguished from the original data, which may indicate lower utility but higher privacy. A lower than 0.5 AA score suggests overfitting, where the synthetic data resemble too closely the original data, potentially compromising privacy.
        \end{itemize}
        
\subsection{Statistical Similarity Metrics} 
\begin{itemize}
    \item \textbf{Wasserstein Distance} quantifies the cost of transforming one distribution into another. The distance can range from 0 to infinity and a smaller value indicates that the two distributions are more similar, while a larger value signifies more substantial differences in the distributions. The Wasserstein-1 distance can be calculated as follows (\ref{eq: w1}):
    \begin{equation}
        W_1(P, Q) = \inf_{\gamma \in \Gamma(P, Q)} \int_{\mathbb{R}^n \times \mathbb{R}^n} \| x - y \| \, d\gamma(x, y)
    \label{eq: w1}
    \end{equation}
    
    \begin{itemize}
        \item $P, Q$: the probability distributions
        \item $\Gamma(P, Q)$: the set of all joint distributions of P and Q
        \item $\| x - y \|$: represents the Euclidean distance between points x and y (in n-dimensions)
        \item $\gamma(x, y)$: joint distributions of x and y that represent the amount of probability mass moved from x to y.
    \end{itemize}
    \item \textbf{Kolmogorov-Smirnov Test}: 
    The KS-Test compares the CDFs of two samples, which can be calculated by (\ref{eq: ks}): 
    
    \begin{equation}
        D_{n,m} = sup_x \left| F_n(x) - G_m(x) \right|
    \label{eq: ks}
    \end{equation}

where:
\begin{itemize}
    \item $D_{n,m}$: the KS test statistic;
    \item $sup_x$: the supremum of the set of absolute differences;
    \item $F_n(x)$: the CDF of the first sample $X$;
    \item $G_m(x)$: the CDF of the second sample $Y$.
\end{itemize}
    
    The KS-test statistic ranges from 0 to 1. A value closer to 0 indicates that the two distributions are similar, while a value closer to 1 suggests significant differences. 
    In our framework, the KS statistic is calculated for each pair of columns in the original and synthetic datasets. To get an overall score at the end for each original-synthetic dataset pair, the mean of the similarity scores is taken (\ref{eq: mean_ks}), where a higher score indicates better similarity:
    \begin{equation}
        \text{Overall Score} = \frac{1}{N} \sum_{i=1}^{N} \left( 1 - D_{n,m}^i \right)
        \label{eq: mean_ks}
    \end{equation}
    \begin{itemize}
        \item $N$: number of columns
        \item $D_{n,m}^i$: the KS-Test statistic for the i-th column.

    \end{itemize}

    \item \textbf{Pearson \& Spearman Correlation:} This metric checks if relationships between variables are preserved. The correlation coefficients range from -1 to 1. A value of 1 indicates a perfect positive correlation, -1 indicates a perfect negative correlation, and 0 means no correlation. For each pair of columns A and B, the correlation coefficients for the original data $O_{A,B}$ and synthetic data $S_{A,B}$ are calculated. To calculate the individual similarity score:

    \begin{equation}
        \text{score}_{A,B} = 1 - \frac{|S_{A,B} - O_{A,B}|}{2}
        \label{eq: ps sim score}
    \end{equation}

    \begin{itemize}
        \item $S_{A,B}$: Correlation coefficient for columns A and B in the synthetic dataset.
        \item $O_{A,B}$: Correlation coefficient for columns A and B in the original dataset.
    \end{itemize}

    The $\text{score}_{A,B}$ ranges from 0 to 1, where a score of 1 indicates that the pairwise correlations of the real and synthetic data are identical and a score of 0 implies that the pairwise correlations are totally different.
    To calculate the final similarity score for correlations between an original and synthetic dataset, the mean of all individual similarity scores is taken:
    \begin{equation}
        \text{Overall Similarity Score} = \frac{1}{N} \sum_{i=1}^{N} \text{score}_i
        \label{eq: ps final}
    \end{equation}
    \begin{itemize}
        \item $N$: Total number of column pairs
        \item $score_i$: Individual similarity scores for each pair.
    \end{itemize}
    
    \item \textbf{Mutual Information}: Pairwise mutual information measures the dependency between two attributes within a dataset. It quantifies how much information knowing one attribute provides about the other. A higher mutual information value indicates a stronger relationship between the two attributes, which means that knowing the value of one attribute reduces the uncertainty about the value of the other. Normalized Mutual Information (NMI) scales the Mutual Information score to lie between 0 and 1, where 0 means no mutual information (independence) and 1 indicates perfect correlation. The calculation of the NMI depends on the average method chosen for normalization, in this case, the arithmetic mean. The formula for NMI is given by (\ref{eq: nmi}):
    \begin{equation}
        \text{NMI}(X, Y) = \frac{2 \cdot MI(X; Y)}{H(X) + H(Y)} 
        \label{eq: nmi}
    \end{equation}

    \begin{itemize}
        \item $MI(X; Y)$: the mutual information between X and Y.
        \item $H(X), H(Y)$: the entropies of X and Y, respectively.
    \end{itemize}

    To calculate the pairwise mutual information between attributes X and Y within a dataset (\ref{eq: mi}):
    \begin{equation}
        MI(X;Y) = \sum_{y \in Y} \sum_{x \in X} p(x,y) \log \left( \frac{p(x,y)}{p(x) p(y)} \right) 
        \label{eq: mi}
    \end{equation}
    \begin{itemize}
        \item $p(x,y)$: the joint probability distribution of X and Y.
        \item $p(x) p(y)$: the marginal probability distributions of X and Y.
    \end{itemize}
    
The absolute differences between all NMI pairs of the synthetic and original dataset are measured and then, similar to (\ref{eq: mean_ks}), subtracted from 1 to convert it to a similarity score that ranges from 0 to 1 and averaged. 

    \item \textbf{Jensen-Shannon Similarity}:
    \label{item_js_sim}
    The JS-Divergence probability distributions. The divergence ranges from 0 to 1. A value closer to 0 indicates that the two probability distributions are similar, while a value closer to 1 suggests significant differences. For each column, the Jensen-Shannon distance is calculated and 1 - distance is used to create a similarity score, where a higher score indicates better similarity.
    For the overall score, the mean of those scores is calculated.
    \item \textbf{Mean, Median, Variance}: 
    \label{item_basic_stats}
    Basic descriptive statistics to see if averages and variability match. These descriptive statistics don’t have fixed ranges but should be compared directly between the synthetic and real datasets. Similar values indicate that the averages and variability match closely, signifying that the synthetic data accurately reflects the properties of the original data.
    For the overall score, the average is calculated.
\end{itemize}

\subsection{Machine Learning Utility}

\begin{figure}
    \centering
    \includegraphics[width=1\linewidth]{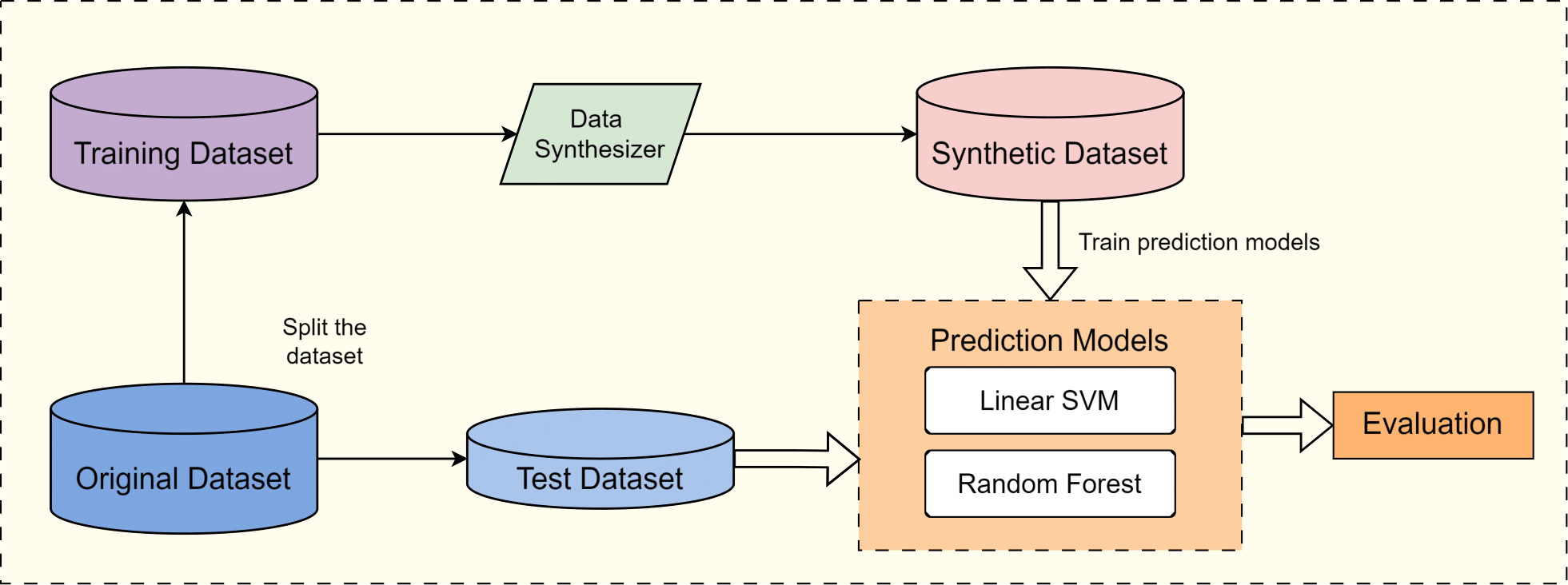}
    \caption{Pipeline for ML Utility Evaluations}
    \label{fig:ml_eval}
\end{figure}

This pillar assesses the utility of synthetic data for machine learning tasks (\emph{cf.} Figure~\ref{fig:ml_eval}). It evaluates the ability of synthetic data to support model training and produce comparable results to models trained on real data. This is essential to ensure that synthetic data can effectively replace real data in ML applications without significant performance degradation.

Overall, this framework provides a comprehensive methodology for evaluating the quality and privacy of synthetic tabular data generation. By combining privacy, statistical similarity, and machine learning utility assessments, it ensures that synthetic data can be effectively and responsibly used in various applications. The proposed framework is implemented in Python and open-sourced, visualization functionalities are also included in the framework.

\section{Evaluation Experiments}
This chapter demonstrates the application of the FEST framework by evaluating six synthetic data generation models across three real-world datasets, serving as a practical use-case and showcasing the framework's capabilities in assessing both the utility and privacy risks associated with synthetic data.






\subsection{Datasets and Generation Models}
Six synthetic data generation models (Section~\ref{chap: gen model}) are evaluated using FEST across three diverse datasets:
\begin{itemize}
    \item Diabetes Dataset: A 768-record dataset derived from the "National Institute of Diabetes and Digestive and Kidney Diseases" dataset \cite{DiabetesDataset}. Features include Pregnancies, Glucose, BloodPressure, SkinThickness, Insulin, BMI, DiabetesPedigreeFunction, Age, and Outcome.
    \item Cardio Dataset: A 70,000-record dataset on cardiovascular disease risk factors \cite{RiskDisease}, containing information on age, gender, height, weight, blood pressure, cholesterol, glucose, smoking, alcohol consumption, physical activity, and cardiovascular disease presence.
    \item Insurance Dataset: A 1,338-record dataset on medical costs \cite{MedicalDatasets}, including age, sex, bmi, children, smoker status, region, and medical costs billed by insurance.
\end{itemize}

\subsection{Results for the Diabetes Dataset}

The privacy assessment run first is followed by the statistical similarity metrics. 

\subsubsection{Privacy Assessment}
The analysis of distance-based privacy metrics \ref{tab:privacy_diabetes} revealed that DiSCO and repU yielded low values for most models (except Random), suggesting strong privacy preservation. However, the choice of appropriate quasi-identifiers and targets for these metrics in this dataset proved challenging. NNDR was highest for CopulaGAN, while DCR was highest for CTGAN. NNAA was closest to the ideal 0.5 for GMM, indicating a good balance between privacy and utility. Attack-based privacy metrics \ref{tab:attack_diabetes} showed that the Random model posed the highest risks, which does make sense because the model simply sampled from the original dataset.

\begin{table}[h]
\centering
\scriptsize
\caption{Comparison of Models Across Different Distance-Based Privacy Metrics for the Diabetes Dataset.}
\label{tab:privacy_diabetes}
\begin{tabular}{lccccc}
\hline
Model & DiSCO & repU & NNDR & DCR & NNAA \\
\hline
CopulaGAN & 0.00 & 0.00 & 0.90 & 0.28 & 0.82 \\
CTGAN & 0.00 & 0.00 & 0.89 & 0.30 & 0.77 \\
GC & 0.00 & 0.00 & 0.89 & 0.21 & 0.72 \\
GM & 0.00 & 0.00 & 0.88 & 0.24 & 0.60 \\
TVAE & 0.00 & 0.00 & 0.86 & 0.16 & 0.70 \\
Random & 100.00 & 100.00 & 0.00 & 0.00 & 0.00 \\
\hline
\end{tabular}
\end{table}

\begin{table*}[h]
    \centering
    \caption{Comparison of Models on Singling Out Risk, Linkability Risk, and Inference Risk with Respective Confidence Intervals for the Diabetes Dataset.}
    \label{tab:attack_diabetes}
    \begin{tabular}{l|ccc}
        \hline
        &S-Out & Link& Inf \\
        \hline
        CopulaGAN&0.1174,CI=(0.0863, 0.1485)&0.0316,CI=(0.0168, 0.0464)&0.1214,CI=(0.0, 0.2710) \\
        CTGAN&0.1347,CI=(0.1024, 0.1670)&0.0336,CI=(0.0182, 0.0489)&0.1781,CI=(0.0, 0.4307) \\
        GC&0.1735,CI=(0.1380, 0.209)&0.0574,CI=(0.0374, 0.0774)&0.4500,CI=(0.2214, 0.6804) \\
        GMM&0.1236, CI=(0.0929, 0.1543)&0.0455,CI=(0.0276, 0.0634)&0.2819,CI=(0.0, 0.5961) \\
        TVAE&0.1662, CI=(0.1293, 0.2031)&0.0514,CI=(0.0325, 0.0704)&0.0858,CI=(0.0, 0.2127) \\
        Random& 0.9961,CI=(0.9923, 1.0)&0.9962,CI=(0.9924, 1.0)&0.9754,CI=(0.9507, 1.0) \\
        \hline
    \end{tabular}
    \label{table:d_risk}
\end{table*}

\subsubsection{Statistical Similarity Assessment}
\begin{figure}
    \centering
    \includegraphics[width=1\linewidth]{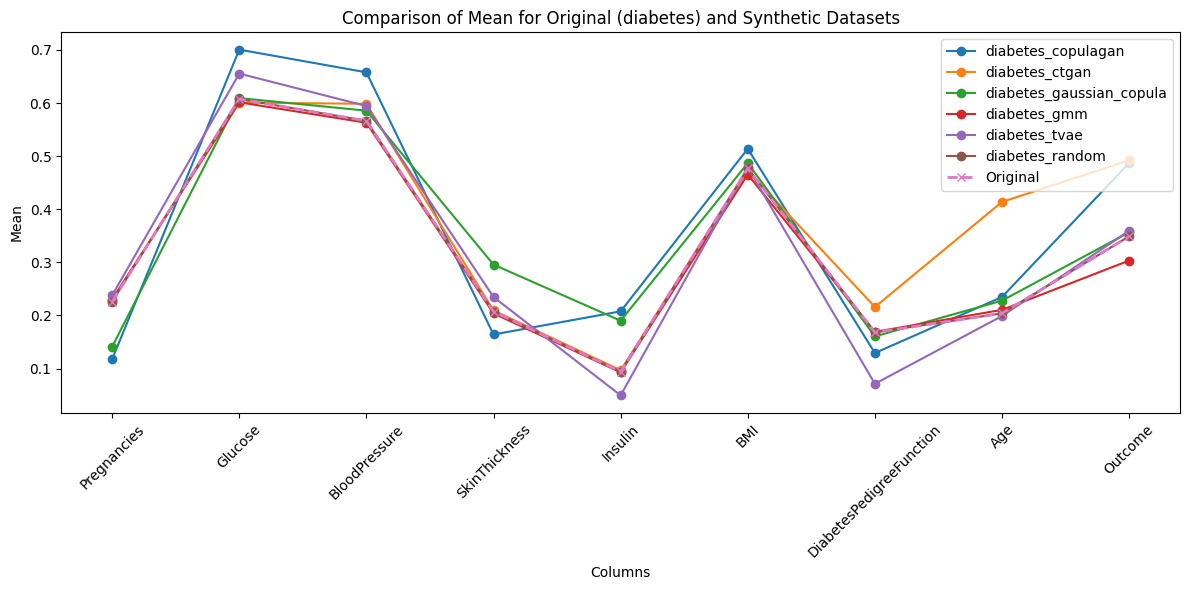}
    \caption{Mean of Diabetes Dataset Compared to Means of Synthetic Datasets.}
    \label{fig:d_basic_stats}
\end{figure}

The synthetic data exhibited high statistical similarity to the original data. This was evident in high KS, P\&S Corr, MI, and JS values. For mean, median, and variance, the difference values are relatively low as well, with TVAE having the lowest (besides random), which measures average distances between real and synthetic statistics. Additionally, plots were made to visualize these means for each column for the original dataset, compared to the means of the columns from all synthetic datasets. An example is shown in \ref{fig:d_basic_stats}.

\begin{table*}[h]
    \centering
        \caption{Comparison of Models Across Different Utility Metrics for the Diabetes Dataset.}
    \begin{tabular}{l|cccccc}
        \hline
        & WS & KS & P\&S Corr & MI & JS & \multicolumn{1}{c}{(Mean, Median, Var)} \\
        \hline
        CopulaGAN & 0.4557 & 0.7477 & [0.9255; 0.9201] & 0.9728 & 0.7818 & (0.0772, 0.0676, 0.0113) \\
        CTGAN & 0.4774 & 0.8422 & [0.9207; 0.9154] & 0.9707 & 0.8130 & (0.0592, 0.0344, 0.0139) \\
        GC & 0.2994 & 0.8556 & [0.9692; 0.9642] & 0.9717 & 0.8277 & (0.0377, 0.0398, 0.0062) \\
        GMM & 0.2814 & 0.9194 & [0.9862; 0.9772] & 0.9617 & 0.8287 & (0.0092, 0.0263, 0.0069) \\
        TVAE & 0.2699 & 0.8588 & [0.9518; 0.9462] & 0.9819 & 0.8440 & (0.0304, 0.0273, 0.0089) \\
        Random & 0.0000 & 1.0000 & [1.0000; 1.0000] & 1.0000 & 1.0000 & (0.0, 0.0, 0.0) \\
        \hline
    \end{tabular}

    \label{tab:d_u}
\end{table*}


\begin{figure}
    \centering
    \includegraphics[width=1\linewidth]{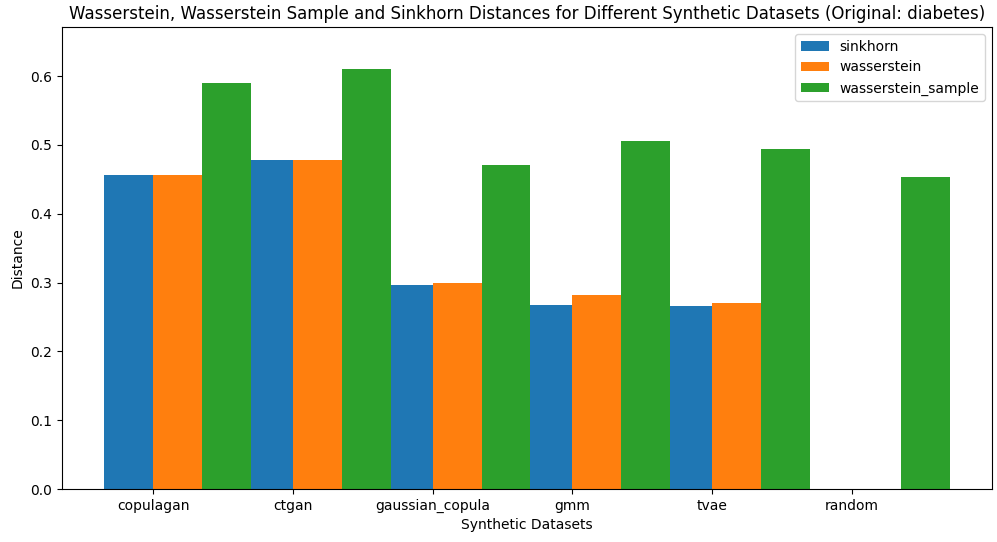}
    \caption{Comparison of Wasserstein Distance Methods (Wasserstein, Wasserstein\_SAMPLE = 20, SINKHORN) for Diabetes Dataset.}
    \label{fig:comparison_Wasserstein}
\end{figure}

\subsection{Results for the Cardio Dataset}

The privacy assessment run first is followed as well by the statistical similarity metrics. 

\subsubsection{Privacy Assessment}
As shown in Table \ref{tab:privacy_cardio}, DiSCO and repU remained low for all models (except Random). For this dataset, the following quasi-identifiers and target was chosen: 
    \texttt{keys = ['age', 'gender', 'height', 'weight', 'cholesterol', 'gluc'], target = "cardio"}. The DCR and NNAA values for this dataset are lower than for the diabetes dataset, which means that the distances between the synthetic and original dataset are smaller and that privacy \& utility are balanced better. This could be due to this dataset being way larger than the diabetes dataset.

\begin{table}[h]
\centering
\scriptsize 
\caption{Comparison of Models Across Different Distance-Based Privacy Metrics for the Cardio Dataset.}
\label{tab:privacy_cardio}
\begin{tabular}{lccccc}
\hline
 & DiSCO & repU & NNDR & DCR & NNAA \\
\hline
CopulaGAN & 0.00 & 0.00 & 0.76 & 0.02 & 0.65 \\
CTGAN & 0.00 & 0.00 & 0.75 & 0.03 & 0.69 \\
GC & 0.00 & 0.00 & 0.80 & 0.03 & 0.64 \\
GMM & 0.00 & 0.00 & 0.81 & 0.16 & 0.64 \\
TVAE & 0.00 & 0.00 & 0.76 & 0.01 & 0.63 \\
Random & 100.00 & 100.00 & 0.00 & 0.00 & 0.00 \\
\hline
\end{tabular}
\end{table}

For the Singling Out risk, 1000 samples were taken from the synthetic and original dataset due to the large number of entries in the cardio dataset, so the risk might not be representative of the actual risk, with no control dataset. But based on the given numbers in \ref{table:c_risk}, the lowest risk was calculated for the GMM.

For the Linkability risk, the train and test datasets were used as calculated by \texttt{dynamic\_train\_test\_split}. Due to the large dataset, 1,000 records were sampled. For the number of attacks, a higher number of 2,000 was chosen, as well as for the number of neighbors as 10, instead of the default value of 1 to achieve a higher number. Attack-based privacy metrics (Table \ref{table:c_risk}) showed generally low risks, except for the Random model.



\begin{table*}[h]
\centering
\caption{Comparison of Models on Singling Out Risk, Linkability Risk, and Inference Risk with Respective Confidence Intervals for the Cardio Dataset.}
\begin{tabular}{l|ccc}
\hline
              & S-Out              & Link                 & Inf                   \\ \hline
CopulaGAN     & 0.1924,CI=(0.158,0.2267)&0.0,CI=(0,0.0021)&0.0511,CI=(0.0293,0.0729) \\
CTGAN         & 0.2182,CI=(0.1822,0.2542)&0.0025,CI=(0,0.0060)&0.0361,CI=(0, 0.0805)    \\ 
GC            & 0.0792,CI=(0.0559,0.1026)&0.0010,CI=(0,0.0050)& 0.0349,CI=(0.0047,0.0651) \\
GMM           & 0.0971,CI=(0.0714,0.1228)&0.0020,CI=(0,0.0054)&0.0141,CI=(0,0.0706) \\ 
TVAE          & 0.2281, CI=(0.1915,0.2647) & 0.0, CI=(0,0.0028)&0.0278, CI=(0,0.0712)    \\ 
Random        & 0.2281, CI=(0.1915,0.2647)&0.1543, CI=(0.1382,0.1705)&0.9994, CI=(0.9988, 1.0) \\ \hline
\end{tabular}

\label{table:c_risk}
\end{table*}

\subsubsection{Statistical Similarity metrics}
\begin{figure}
    \centering
    \includegraphics[width=1\linewidth]{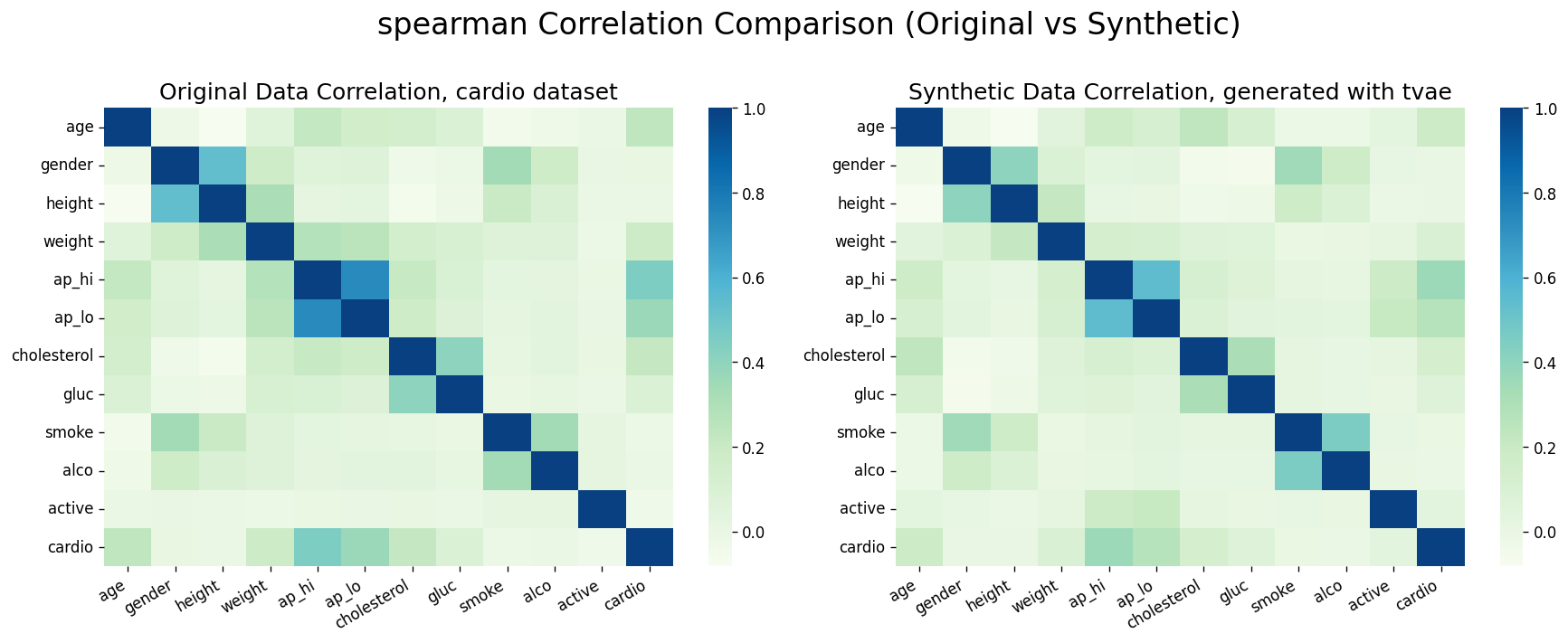}
    \caption{Spearman Correlation Heatmap: Comparison between Cardio Dataset and Synthetic Dataset generated with TVAE.}
    \label{fig:cardio_Spearman_corr}
\end{figure}

For the Wasserstein distance in Table \ref{tab:c_util}, the Sinkhorn approximation was used because the dataset was too large and complex. All other metrics were computed normally. For the other metrics KS, P\&S Corr, MI and JS, this study can see that the results are again high, close to 1 which supports statistical similarity. As a visualization, the heatmap of the Spearman correlation of both the synthetic (generated with TVAE) and original dataset was included Figure \ref{fig:cardio_Spearman_corr}. The correlations are quite comparable, though there are slight variations. For instance, certain regions of the synthetic data heatmap appear lighter, indicating a reduced correlation between the columns.

For this dataset, the values in the (Mean, Median, Var) column are closer to zero compared to the diabetes dataset. This might be attributed to having a larger amount of data available to produce synthetic samples that match the original distribution, along with having more records overall.

\begin{table*}[h]
    \centering
        \caption{Comparison of Models Across Different Utility Metrics for the Cardio Dataset.}
    \begin{tabular}{l|cccccc}
    \hline
    & WS & KS & P\&S Corr & MI & JS & (Mean, Median, Var) \\ \hline
    CopulaGAN & 0.2758 & 0.9479 & [0.9598, 0.9602] & 0.9889 & 0.8888 & (0.0197, 0.0886, 0.0093) \\     CTGAN & 0.4430 & 0.8865 & [0.9716, 0.972] & 0.9841 & 0.7784 & (0.0465, 0.0075, 0.0254) \\ 
    GaussianCopula & 0.2461 & 0.9321 & [0.9734, 0.9683] & 0.9779 & 0.8179 & (0.0007, 0.0846, 0.0005) \\ 
    GMM & 0.3037 & 0.9369 & [0.996, 0.9921] & 0.973 & 0.7758 & (0.0025, 0.0861, 0.0223) \\ 
    TVAE & 0.2204 & 0.9412 & [0.9805, 0.9785] & 0.9861 & 0.8476 & (0.0158, 0.0859, 0.0095) \\ 
    Random & 0.1777 & 1.0000 & {[1.0000:1.0000]} & 1.0000 & 1.0000 & (0.0, 0.0, 0.0) \\ \hline
    \end{tabular}

    \label{tab:c_util}
\end{table*}

\subsection{Results for the Insurance Dataset}

The privacy assessment run first is followed again by the statistical similarity metrics. 

\subsubsection{Privacy Assessment}

As can be seen in \ref{tab:privacy_insurance}, the values of DiSCO are again almost all 0, which raises the question if the quasi-identifiers and the target were chosen in a wrong way, or if the DiSCO is unsuitable for such kind of datasets. The quasi-identifiers were chosen as \texttt{keys = ['age', 'bmi', 'children']} with \texttt{target = charges}. The insurance dataset may be of medium size but does not have as many attributes to find appropriate quasi-identifiers. Users can choose their own quasi-identifier with the proposed framework. The repU was highest for TVAE. NNDR, DCR, and NNAA results were comparable to the other datasets. For the NNDR, DCR and NNAA, the results are very similar as in the other two datasets.

\begin{table}[h]
\centering
\scriptsize 
\caption{Comparison of Models Across Different Distance-Based Privacy Metrics for the Insurance Dataset.}
\label{tab:privacy_insurance}
\begin{tabular}{lccccc}
\hline
 & DiSCO & repU & NNDR & DCR & NNAA \\
\hline
CopulaGAN & 0.00 & 0.07 & 0.84 & 0.28 & 0.75 \\
CTGAN & 0.00 & 0.00 & 0.83 & 0.27 & 0.73 \\
GC & 0.00 & 0.07 & 0.79 & 0.17 & 0.62 \\
GMM & 0.00 & 0.00 & 0.77 & 0.14 & 0.58 \\
TVAE & 0.00 & 0.15 & 0.75 & 0.11 & 0.69 \\
Random & 98.65 & 98.51 & 0.00 & 0.00 & 0.00 \\
\hline
\end{tabular}
\end{table}

All singling out risks are around 0.1-0.15, except for the random model having a risk of basically 1. For the linkability risk, the following auxiliary columns were chosen:
\texttt{aux\_cols\_i = (["age", "sex", "bmi"], ["children", "smoker", "region", "charges"])}.
In this calculation, the number of attacks had to be lowered to 260 due to the test dataset being smaller than the default value of 500. The number of neighbors was chosen as 10, the number of attacks as 2000. The final risks are really low, which raises the question of whether the linkability risk can be accurately measured for this dataset and selection of parameters.

\begin{table*}[h]
    \centering
    \caption{Comparison of Models on Singling Out Risk, Linkability Risk, and Inference Risk with Respective Confidence Intervals for the Insurance Dataset.}
    \begin{tabular}{l|ccc}
        \hline
        & S-Out & Link & Inf \\ \hline
        CopulaGAN & 0.1249, CI=(0.0962, 0.1536) & 0.0128, CI=(0.0, 0.073) & 0.0414, CI=(0.0, 0.1957) \\
        CTGAN     & 0.1090, CI=(0.0820, 0.1361) & 0.0, CI=(0.0, 0.011) & 0.02042, CI=(0.0, 0.2415) \\
        GC        & 0.1566, CI=(0.1250, 0.1883) & 0.0, CI=(0.0, 0.0080) & 0.0588, CI=(0.0, 0.1252) \\
        GMM       & 0.1011, CI=(0.0749, 0.1272) & 0.01501, CI=(0.0, 0.1113) & 0.1025, CI=(0.0069, 0.1980) \\
        TVAE      & 0.1229, CI=(0.0944, 0.1514) & 0.0045, CI=(0.0, 0.0784) & 0.2732, CI=(0.0, 0.5786) \\
        Random    & 0.9962, CI=(0.9924, 1.0) & 0.9890, CI=(0.9780, 1.0) & 0.9907, CI=(0.9813, 1.0) \\ \hline
    \end{tabular}
    \label{tab:i_risk}
\end{table*}

\subsubsection{Statistical Similarity Assessment}
\begin{figure}
    \centering
    \includegraphics[width=1\linewidth]{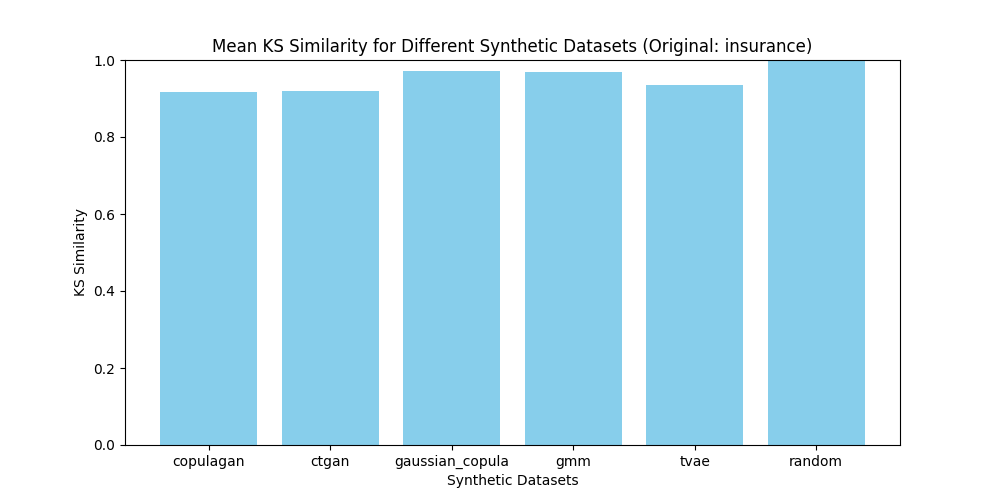}
    \caption{Barplot of the KS-Similarity Scores of Different Synthetic Datasets.}
    \label{fig:i_mean_ks}
\end{figure}

\begin{figure}
    \centering
    \includegraphics[width=1\linewidth]{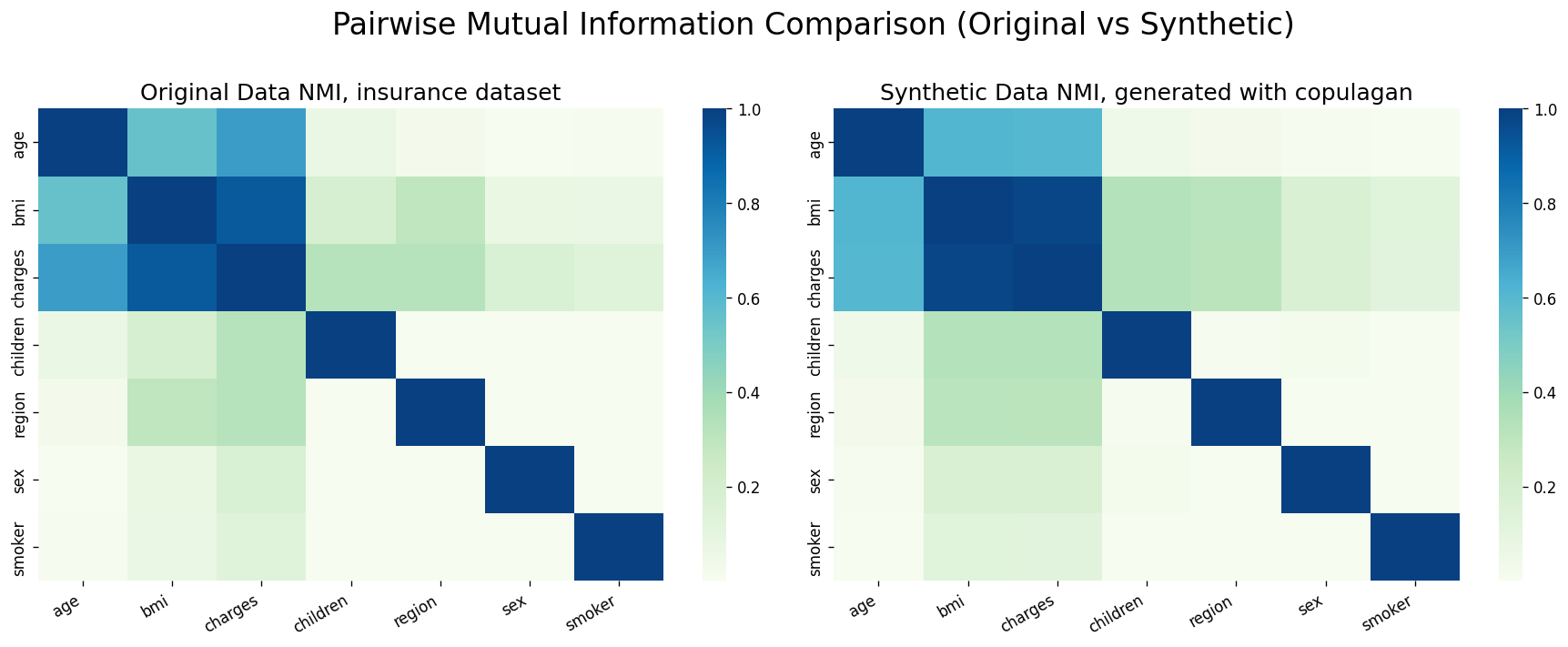}
    \caption{Comparison of NMIs of Insurance and Synthetic Dataset Generated with CopulaGAN}
    \label{fig:i_mi}
\end{figure}

Two plots, Figure \ref{fig:i_mean_ks} and \ref{fig:i_mi}, were provided as examples for the results found in Table \ref{tab:i_util}. The WS of Table \ref{tab:i_util} has higher values compared to Table \ref{tab:c_util}, but similar scores for all other columns. The KS similarity scores are visualized as a bar graph, with the Gaussian copula having the highest overall score with 0.9727, after the Random Model with 1.0. Another example, depicted in \ref{fig:i_mi}, shows the heatmap of the normalized mutual information outcomes of the insurance and the synthetic dataset generated with CopulaGAN. The finer regions seen in the heatmap on the left are less detailed compared to those on the right.


\begin{table*}[h]
    \centering
        \caption{Comparison of Models Across Different Utility Metrics for the Insurance Dataset.}
    
    \begin{tabular}{l|cccccc}
    \hline
    & WS & KS & P\&S Corr & MI & JS & \multicolumn{1}{c}{(Mean, Median, Var)} \\ \hline
    CopulaGAN & 0.4444 & 0.9174 & {[}0.9714, 0.9711{]} & 0.9874 & 0.9126 & (0.0449, 0.0338, 0.0181) \\ 
    CTGAN & 0.4141 & 0.9207 & {[}0.9676, 0.969{]} & 0.988 & 0.9236 & (0.0415, 0.1931, 0.0136) \\ 
    GaussianCopula & 0.2939 & 0.9727 & {[}0.9742, 0.9783{]} & 0.9877 & 0.9666 & (0.0158, 0.0047, 0.006) \\ 
    GMM & 0.2450 & 0.9682 & {[}0.9906, 0.9888{]} & 0.985 & 0.9353 & (0.0104, 0.0035, 0.0034) \\ 
    TVAE & 0.4691 & 0.9340 & {[}0.9554, 0.9575{]} & 0.9831 & 0.9341 & (0.0349, 0.0295, 0.0136) \\ 
    Random & 0.0000 & 1.0000 & {[}1.0000;1.0000{]} & 1.0000 & 1.0000 & (0.0, 0.0, 0.0) \\ \hline
    \end{tabular}

    \label{tab:i_util}
\end{table*}

\subsection{Summary and Discussion of the Experiments}

The FEST framework demonstrated several key advantages:

\begin{itemize}
    \item Comprehensiveness: It encompasses a wide range of privacy and utility metrics, providing a holistic evaluation.
    \item Flexibility: The framework can be adapted to different datasets and evaluation scenarios by adjusting parameters and incorporating new metrics.
    \item Interpretability: The results from FEST are relatively easy to interpret and compare across models, aiding in model selection and refinement.
\end{itemize}

The results provide insights into the privacy and utility trade-offs associated with different models and datasets. The framework's flexibility and comprehensiveness make it a valuable tool for researchers and practitioners in the field of synthetic data generation. The evaluation also highlighted the limitations of specific metrics, such as DiSCO and repU, particularly when dealing with smaller datasets or in the absence of suitable quasi-identifiers. Future research will focus on refining these metrics and expanding the framework to include additional evaluation techniques.

\section{\uppercase{Conclusions}}
\label{sec:conclusion}

This research presents FEST, a framework for comprehensively evaluating the privacy and utility of synthetic tabular data. FEST addresses the limitations of existing evaluation methods by integrating a diverse set of privacy metrics, including distance-based measures (DiSCO, repU, NNDR, DCR, NNAA) and attack-based metrics (Singling Out Risk, Linkability Risk, Inference Risk). These are combined with utility metrics such as Wasserstein Distance, KS test, Pearson and Spearman Correlation, Mutual Information, and Jensen-Shannon Similarity to provide a holistic assessment of synthetic data quality.

Through the evaluation on three diverse datasets (Diabetes, Cardio, and Insurance), FEST demonstrated its effectiveness in analyzing the trade-offs between privacy and utility for various synthetic data generation models, including CopulaGAN, CTGAN, GaussianCopula, GMM, TVAE, and a Random baseline. The results highlighted the importance of considering a wide range of metrics to gain a comprehensive understanding of the strengths and weaknesses of different synthetic data generation techniques. 

By providing a standardized and flexible framework, FEST empowers researchers and practitioners to make informed decisions about the suitability of synthetic data for specific applications.




\bibliographystyle{apalike}
{\small
\bibliography{reference}}

\section*{\uppercase{Acknowledgments}}
This work was partially supported by \emph{(a)} the University of Zürich UZH, Switzerland, and \emph{(b)} the Horizon Europe Framework Program's project AISym4MED, Grant Agreement No.101095387, funded by the Swiss State Secretariat for Education, Research, and Innovation SERI, under Contract No.22.00622.

\end{document}